\documentclass{article} % For LaTeX2e
\usepackage{collas2022_conference,times}
\usepackage{float}
\usepackage{caption}    
\usepackage{subfig}   
\usepackage{graphicx}
\usepackage{amsmath}

\usepackage{amsmath,amsfonts,bm}

\def\eqref#1{equation~\ref{#1}}
\def\1{\bm{1}}

\DeclareMathAlphabet{\mathsfit}{\encodingdefault}{\sfdefault}{m}{sl}
\SetMathAlphabet{\mathsfit}{bold}{\encodingdefault}{\sfdefault}{bx}{n}

\usepackage{hyperref}
\hypersetup{
    colorlinks=true,
    linkcolor=red,
    filecolor=magenta,      
    urlcolor=blue,
    citecolor=purple,
    pdftitle={Overleaf Example},
    pdfpagemode=FullScreen,
    }

\title{Feature diversity in self-supervised learning}

\newcommand*\samethanks[1][\value{footnote}]{\footnotemark[#1]}

\author{Pranshu Malviya \thanks{Equal contribution} \\
Department of Computer and Software Engineering\\
Mila - Quebec AI Institute\\
École Polytechnique de Montréal\\
Montréal, Canada \\
\And % Use And to have authors side by side
Arjun Vaithilingam Sudhakar \samethanks[1]\\
Department of Computer Science and Operations Research \\
Mila - Quebec AI Institute \\
Université de Montréal \\
Montréal, Canada \\
}

\collasfinalcopy % Uncomment for camera-ready version, but NOT for submission.

\begin{document}

\maketitle

\begin{abstract}
Many studies on scaling laws consider basic factors such as model size, model shape, dataset size, and compute power. These factors are easily tunable and represent the fundamental elements of any machine learning setup. But researchers have also employed more complex factors to estimate the test error and generalization performance with high predictability. These factors are generally specific to the domain or application. For example, feature diversity was primarily used for promoting syn-to-real transfer by \cite{chen2021contrastive}. With numerous scaling factors defined in previous works, it would be interesting to investigate how these factors may affect overall generalization performance in the context of self-supervised learning with CNN models. How do individual factors promote generalization, which includes varying depth, width, or the number of training epochs with early stopping? For example, does higher feature diversity result in higher accuracy held in complex settings other than a syn-to-real transfer? How do these factors depend on each other? We found that the last layer is the most diversified throughout the training. However, while the model's test error decreases with increasing epochs, its diversity drops. We also discovered that diversity is directly related to model width.
\end{abstract}

\section{Introduction}
In recent years, self-supervised learning (SSL) has achieved exceeding empirical success \citep{DBLP:journals/corr/abs-1911-05722} and also a method of pretraining neural networks \citep{DBLP:journals/corr/abs-2006-09882}. A major advantage of SSL as compared to supervised learning is the ability to scale since SSL requires no manual labeling process. \cite{DBLP:journals/corr/scaling_benchmark_ssl_visual} scales the dataset and difficulty of the problem for this study. They found that the results matched and even surpassed those of supervised learning techniques. They also released a benchmark for $9$ different datasets and tasks for evaluation. Once trained, these models can be used to learn new tasks more data-efficiently by finetuning \citep{chen2021contrastive}.

Contrastive learning is a type of SSL technique that pulls representations of the anchor image and its transformations closer and pushes the images from the different classes farther. \cite{survey_contrastive_ssl} and \cite{DBLP:journals/corr/contrastive_rep_learning_review} discuss about contrastive learning's superior performance and the inductive bias of self-supervised algorithms.

With the growing interest in techniques like contrastive learning and invariant predictions \citep{mitrovic2020representation}, it is imperative to discover the factors promoting generalization and scaling laws of state-of-the-art CNN models under the SSL setup \citep{vision_models_robust_fair_facebook}.

Generalization plays a key role to measure the performance of the model. The importance of the role of diversity of learned feature embedding in terms of generalization is studied by \cite{chen2021contrastive} and \cite{ DBLP:journals/corr/hyperspherical}. The lack of diversity in the representation learned by the model makes the prediction sensitive to natural fluctuations in the real world. Hence, it is essential to understand what factors promote generalization in the CNN architecture. We investigate whether increased feature diversity leads to improved accuracy and generalization in complex self-supervised algorithms.

\cite{DBLP:journals/corr/hyperspherical} regularized the neural network by minimum hyperspherical energy (MHE) in order to avoid undesired representation because of the over-parametrization. Also, \cite{pmlr-v70-xie17b} proposed a regularization based on the uncorrelations and evenness that promotes diversity. This will promote the components to be uncorrelated and to have equal roles in data modeling.

Diversity in representation makes the learned self-supervised models resilient to natural variations in the real world. Hence, in this work, we investigate how specific characteristics like depth, width, or the number of training epochs attribute generalization through diversity. We also perform experiments to analyze how test loss relates to the diversity metric. Does higher feature diversity result in better performance? In order to estimate the diversity of each component, we use hyperspherical potential energy.

\section{Approach}
\label{headings}

\subsection{Self-Supervised Learning}

Self-supervised learning receives supervisory signals from the data itself, mostly utilizing the data's underlying structure \citep{DBLP:journals/corr/abs-2006-08218}. SimCLR \citep{DBLP:journals/corr/abs-2002-05709} is an approach based on contrastive learning to learn the visual representation. The representations are learned from the input data by maximizing the agreement between augmented images of the same image through contrastive loss in the latent space.

The loss function for the SimCLR is defined as,
\[L_{SimCLR}^{(i,j)} = -log\frac{\text{exp}(\text{sim}(z_i,z_j)/\tau)}{\sum_{k=1}^{2N} \mathbf{1}_{[k \ne i]} \text{exp}(\text{sim}(z_i,z_k)/\tau)} \label{simclr_loss}
\]
where, $z_i=g(h_i)$ , $z_j=g(h_j)$ and $\mathbf{1}_{[k \ne i]}$ acts as an indicator function, returning 1 if $k \ne i$ is present , and 0 otherwise. In the above equation, $\text{sim}(.)$ represents the cosine similarity. This loss is based on $g(.)$  representation's extra projection layer. Also, $h$ representation is used only for the downstream task. 

We also used other self-supervised learning algorithms like DeepCluster-V2 \citep{DBLP:journals/corr/abs-1807-05520} which acts as an end-to-end system where the parameters of the network and the clustering assignments of the features are learned. In JigSaw \citep{DBLP:journals/corr/NorooziF16} the pretext task to learn the representations is jigsaw puzzles. (tiles are taken from the images and shuffled). We also used RotNet \citep{gidaris2018unsupervised}, which learns the image representation by using CNN to predict the 2D rotations of the input images. This approach will help the model learn the semantic information in the image without any labeled data. In NPID \citep{Wu_2018_CVPR} (Non-Parametric Instance Discrimination) is also a self-supervised algorithm that is based on the non-parametric classification approach. Finally, PIRL \citep{DBLP:journals/corr/abs-1912-01991} based on the pretext task, the invariant representations are learnt. The most commonly used pretext task is solving jigsaw puzzles.

\subsection{Estimating Feature Diversity by Minimum Hyperspherical Energy}
Generally, to handle large datasets, we use large neural networks, which will offer the capacity to fit the data using complex functions. This high degree of representation helps to perform difficult tasks but sometimes results in highly correlated neurons, which can impair generalization ability and incur extra computing costs. To understand the diversity of the features learned by each component of the Resnet architecture, we take motivation from \cite{DBLP:journals/corr/abs-1805-09298} to get a quantitative measure.

The Hyperspherical potential energy will provide us the measure of diversity in the feature embeddings \citep{DBLP:journals/corr/abs-2104-02290}.
%$$E_s(\hat{v}_i|_{i=1}^{N}) = \sum_{i=1}^{N} \sum_{j=1,j\ne i }^{N} e_s(||\hat{v}_i - \hat{v}_j||) =
\[
E_s(\hat{v}_i|_{i=1}^{N}) = \sum_{i=1}^{N} \sum_{j=1,j\ne i }^{N} e_s(||\hat{v}_i - \hat{v}_j||) \rvert = \left\{
        \begin{array}{ll}
            \sum_{i \ne j} ||\hat{v}_i - \hat{v}_j||^{-s},  & \quad s > 0 \\
            \sum_{i \ne j} log(||\hat{v}_i - \hat{v}_j||^{-1}),  & \quad s = 0
        \end{array}
    \right. \label{diversity_eq}
 \]
In the above equation, N represents the number of examples, $||.||$ represents the Euclidean distance, $f_{s}(.)$ is a function of decreasing real value. $\hat{v}$ represents the $i^{th}$ neuron weight projection into the unit hypersphere. Also, here $s$ represents the power factor.  The lowers $E_s$ (Hyperspherical energy) means the feature vectors are more diverse and scattered onto the unit sphere.

% \section{Model and Experiments}

% This section is a mix of models and experiments. This is the main section of the report which will log the entire research process. You will write down the precise description of the model, experiment design, results, graphs, analysis of results, next set of experiments, and so on.

% As the model evolves over time, do not edit the already described model. Write a new version of the model. Treat this section as a research log. Not as a research paper.

% Use the images folder to put all your images. Use \textit{meetings.tex} to maintain a log of all the project meetings and weekly to-do list.
\section{Experiments}

In this section, we describe the experiments performed to analyze the features diversity in models trained using SSL algorithms. These experiments are performed on ImageNet \citep{imagenet_cvpr09} where we train models for larger training epochs, different CNN architectures, and SSL algorithms.

We use different versions of ResNet models with the different number of parameters. The models were based on different choices of CNN components: depth ($50, 101, 152, 200$) and width ($1, 2, 4$). %, head size ($256, 512, 2048$). % and norm (\textit{Batch-norm} and \textit{Layer-norm}). 
The implementation of the base ResNet models is based on the VISSL library \citep{vissl}. The choices of hyper-parameters are also based on default configurations used by \cite{vissl}.

Once the given model is trained using a SSL algorithm, we randomly sample $500$ images from held-out test data and forward-pass to extract the features of the following layers: \textbf{(i)} First convolutional layer (\textit{conv1}), \textbf{(ii)} Second residual block (\textit{res2}), \textbf{(iii)} Forth residual block (\textit{res4}) and \textbf{(iv)} Embedding layer (\textit{Head}). These extracted features are then used to compute feature diversity using hyperspherical energy $E_s^{(l)}$ (where $l$ denotes the layer) defined in Eq. \ref{diversity_eq}. Next, we load the trunk of this trained model and add a linear classification layer at the end. We train this classification layer using labeled data and test the performance on a held-out dataset. We record error ($=1-a_1$, where $a_1$ is the top-1 accuracy) obtained on the test data. In our results, we compare test error and diversity ($= - E_s^{(l)}$) for different ResNet architectures and training regimes. 

% \subsection{ImageNet}
% In this section, we describe the large-scale experiments that were performed on the ImageNet dataset. Similar to small-scale experiments, 

We analyze the characteristics of different layers of the model with \textit{Batch-norm} and train a ResNet model using \textit{SimCLR} with varying epochs, depth, and width. By comparing the layer-wise feature diversity, we could infer whether diversity always results in better performance. We also report the results from small-scale experiments on  CIFAR100 \citep{krizhevsky2009learning} which are performed with a lower number of training epochs for different shapes of the model in Appendix \ref{appendix}.
\begin{figure}[t!]
    \centering
    % % \begin{subfigure}
        \includegraphics[width=0.48\textwidth]{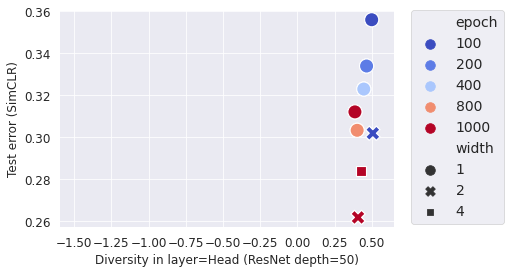}
    % % \end{subfigure}
    % \hfill
    % % \begin{subfigure}
        \includegraphics[width=0.48\textwidth]{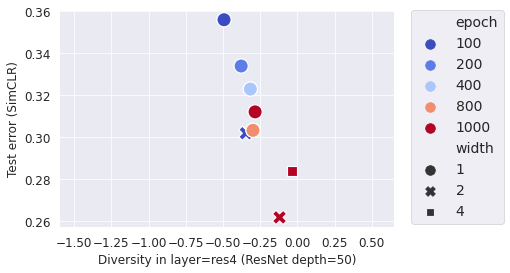}
    % % \end{subfigure}
    \\
    % % \begin{subfigure}
        \includegraphics[width=0.48\textwidth]{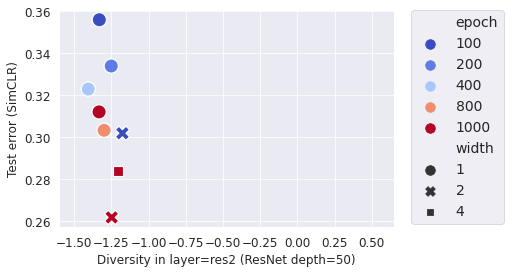}
    % % \end{subfigure}
    % \hfill
    % % \begin{subfigure}
        \includegraphics[width=0.45\textwidth]{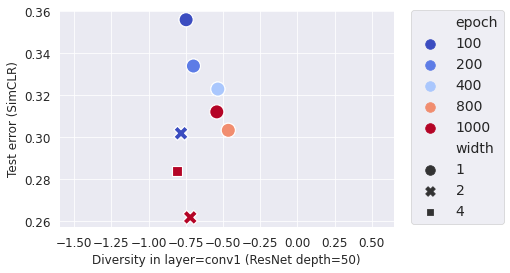}
    % % \end{subfigure}
    \caption{Test error vs feature diversity for varying number of epochs for ResNet models with $depth=50$ trained using \textit{SimCLR}.}
    \label{epoch_fig}
\end{figure}
\subsection{Epochs}
We start by comparing feature diversity with test error for a varying number of epochs by fixing the depth of the model to $50$ and training it using \textit{SimCLR}. The goal of this experiment is to analyze the evolution of feature diversity with increasing epochs and whether early-stopping results in a model with the most diverse features. We also vary the width of the model and plot corresponding results in Figure \ref{epoch_fig}. 

We observe that \textit{Head} layer has the highest feature diversity as compared to other layers. There is also an increasing trend among layers from \textit{res2} to \textit{Head} layer. But the features in \textit{conv1} layer still appear to be more diverse than \textit{res2}. Apart from that, we observe that with increasing width, feature diversity improves in all layers except the first layer i.e., \textit{conv1}. This could be because, with wider hidden layers, the knowledge learned by the model is more spread out among the hidden layers. On the other hand with $width=1$, the first layer learns relatively more diverse features to achieve better performance. In fact, \textit{conv1} features diversity in $width=1$ model increases even after the model is overfitted (at $1000$ epochs). We also observe a correlation between diversity and performance in all layers except \textit{Head} layer. In particular, the model with $width=1$ achieves minimum test error after training for $800$ epochs, whereas the feature diversity in the final layer maximum during $100$ epochs. A gradual decrease in feature diversity of the final layer suggests that there could be a trade-off between classification performance and feature diversity of the model in the final layer.

\begin{figure}[h!]
    \centering
    % \begin{subfigure}
        \includegraphics[width=0.48\textwidth]{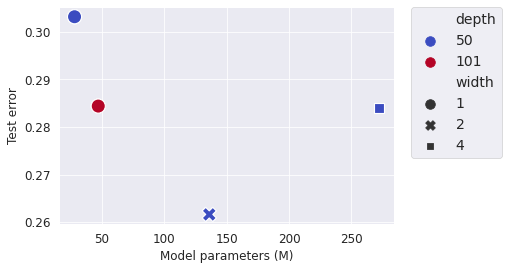}
    % \end{subfigure}
    % \hfill
    % \begin{subfigure}
        \includegraphics[width=0.48\textwidth]{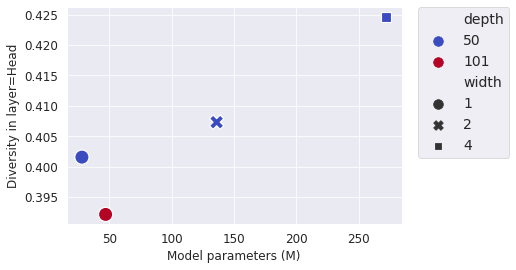}
    % \end{subfigure}
    \caption{Test error and diversity vs. the number of parameters in the models trained using \textit{SimCLR}.}
    \label{simclr_fig}
\end{figure}

Next, we add early-stopping criteria and plot test error and diversity across different model sizes in Figure \ref{simclr_fig} trained using \textit{SimCLR}. For an increasing number of parameters in the model, we observe that test error doesn't always decrease. We also note that diversity improves with an increasing number of parameters, especially in the wider networks. 

\subsection{Algorithms}
\begin{figure}[h!]
    \centering
    % % \begin{subfigure}
        \includegraphics[width=0.8\textwidth]{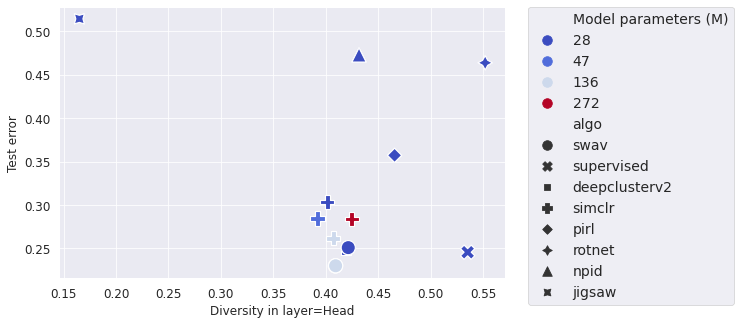}
    % % \end{subfigure}
    \caption{Test error vs. feature diversity for models trained using different SSL algorithms.}
    \label{algo_fig}
\end{figure}

Next, we compare test error and feature diversity for models which are trained using different SSL algorithms. We also compare the performances of these SSL algorithms with the supervised setting in Figure \ref{algo_fig}. We observe that for constant model size, supervised training results in the best feature diversity with minimum test error. We also observe that \textit{DeepClusterV2} results in the best test error as compared to other SSL algorithms for the same model size. \textit{Rotnet} results in best feature diversity but results in high test error.

\section{Conclusion}
We provide a brief survey of different SSL algorithms and the importance of feature diversity in improving feature diversity. When a model is trained using an SSL algorithm, the key idea was to examine how feature diversity for various model layers behaves as compared to the classification error. We performed different experiments on the ImageNet dataset by constructing models with different architectures with varying depth and width. We also vary the number of training epochs to check whether applying early-stopping results in a model with the most diverse features. We found that the final layer remains the most diverse layer throughout the training regime. But although the model's test error decreases, its diversity also decreases with increasing epochs. We also found that diversity is directly proportional to the width of the model. Overall, understanding the behavior of diversity in final layer features and exploiting layer-wise diversity to improve generalization pose interesting directions for future research.

\bibliography{collas2022_conference}
\bibliographystyle{collas2022_conference}

\appendix
\section{Appendix}\label{appendix}

% \subsection{CIFAR100}
The goal of the CIFAR100 experiment was to analyze the characteristics of different layers of the model during the initial stages of the training process. By comparing the layer-wise feature diversity, we could infer whether diversity always results in better performance. 
% With a smaller number of training epochs, we could compare the above metrics across a vast number of architecture choices (with a different number of parameters). 

\begin{figure}[t!]
    \centering
    % \begin{subfigure}
        \includegraphics[width=0.8\textwidth]{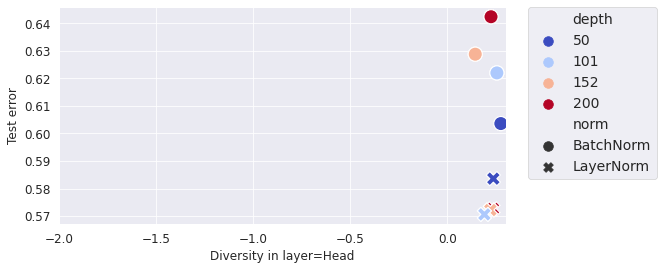}
    % \end{subfigure}
    \\
    % \begin{subfigure}
        \includegraphics[width=0.8\textwidth]{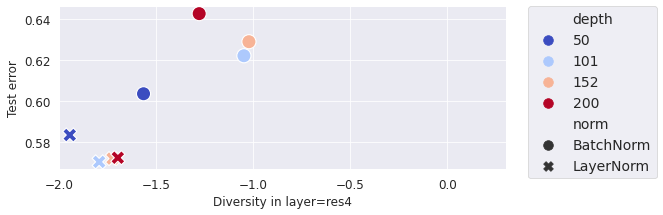}
    % \end{subfigure}
    \\
    % \begin{subfigure}
        \includegraphics[width=0.8\textwidth]{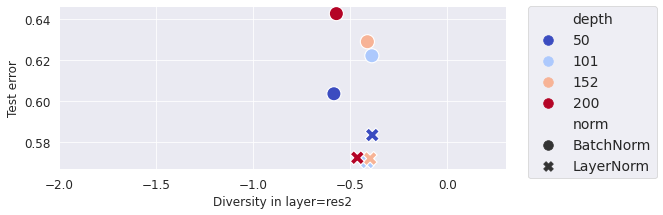}
    % \end{subfigure}
    \\
    % \begin{subfigure}
        \includegraphics[width=0.8\textwidth]{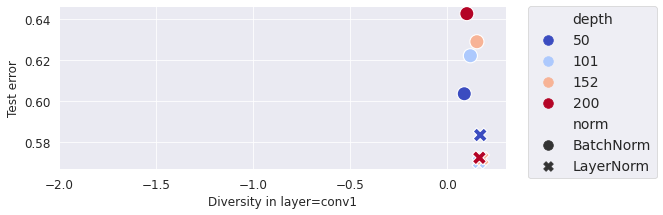}
    % \end{subfigure}
    \caption{Test error vs feature diversity of different layers of the models (with varying depth and choice of norms) during initial stages of the training process.}
    \label{cifar_fig}
\end{figure}

This experiment was performed on the CIFAR100 dataset \citep{krizhevsky2009learning}. We train a ResNet model for $30$ epochs with varying depth sizes and choices of norms using \textit{SimCLR} loss function. We plot the results in Figure \ref{cifar_fig}. We observe that \textit{Head} is the most diverse layer as compared to any other layers in the ResNet. Moreover, none of the layers indicate the correlation between diversity and the performance of the model. Second to \textit{Head} layer, it is the \textit{conv1} layer that is most diverse during initial stages of \textit{SimCLR} learning. We can also see a decreasing trend in feature diversity from \textit{conv1} to \textit{res4} layers. Apart from that, a high test error with increasing depth suggests that the deeper model requires more training epochs for better performance. But even with fewer epochs, we observe that deeper models can learn diverse features. We also observe a greater diversity and lower test error when using LayerNorm instead of BatchNorm. % We perform a similar experiment on the ImageNet dataset and present our analysis in Appendix \ref{imagenet_small}.

\end{document}